# Artificial Intelligence-assisted Biomedical Literature Knowledge Synthesis to Support Decision-making in Precision Oncology

Ting He, MS[1,2], Kory Kreimeyer, MS[1,2], Mimi Najjar, MD[1,3], Jonathan Spiker, BS[1,2], Maria Fatteh, MD[1,3], Valsamo Anagnostou, MD, PhD[1,3], Taxiarchis Botsis, MS, MPS, PhD[1,2]

[1]Sidney Kimmel Comprehensive Cancer Center, Johns Hopkins School of Medicine, Baltimore, MD; [2]Division of Quantitative Sciences, Sidney Kimmel Comprehensive Cancer Center, Johns Hopkins School of Medicine, Baltimore, MD; [3]The Johns Hopkins Molecular Tumor Board, Johns Hopkins School of Medicine, Baltimore, MD

## Abstract

*The delivery of effective targeted therapies requires comprehensive analyses of the molecular profiling of tumors and matching with clinical phenotypes in the context of existing knowledge described in biomedical literature, registries, and knowledge bases. We evaluated the performance of natural language processing (NLP) approaches in supporting knowledge retrieval and synthesis from the biomedical literature. We tested PubTator 3.0, Bidirectional Encoder Representations from Transformers (BERT), and Large Language Models (LLMs) and evaluated their ability to support named entity recognition (NER) and relation extraction (RE) from biomedical texts. PubTator 3.0 and the BioBERT model performed best in the NER task (best F1-score 0.93 and 0.89, respectively), while BioBERT outperformed all other solutions in the RE task (best F1-score 0.79) and a specific use case it was applied to by recognizing nearly all entity mentions and most of the relations. Our findings support the use of AI-assisted approaches in facilitating precision oncology decision-making.*

## Introduction

Precision oncology aims at delivering patient-tailored cancer therapies by targeting specific molecular alterations after in-depth characterization of the molecular profiling of tumors[1]. This task requires thoroughly reviewing several sources, including electronic health records and next-generation sequencing outputs, to accurately define the patient's clinical-genomic phenotype and information from biomedical literature, several knowledge bases, clinical guidelines, clinical trials, FDA-approved and off-label treatments, and more. In practice, medical experts often collaborate in Molecular Tumor Boards (MTB) to interpret patients' clinical and molecular profiles and suggest genotype-targeted therapies based on levels of evidence retrieved and ranked from biomedical resources. This process typically demands manual and labor-intensive steps as existing automated approaches are limited and often do not meet the expected level of performance required to assist human experts. Considering the increasing volume of knowledge generated in all the above sources and the evolving landscape of precision oncology, significant automation is necessary to streamline information extraction from various sources, such as biomedical literature, and combine it with existing and new knowledge to support expert review in the MTB and other use cases.

Automated information retrieval from biomedical texts is generally supported by NLP techniques, and numerous studies have been conducted on this end over the last decades. The exciting capabilities of LLMs in language understanding and generation open new opportunities for research in the domain, and they have already been applied to clinical text for NER purposes[2]. Published literature is a major knowledge source in biomedicine, and several researchers have used NLP and advanced methods to process large corpora of published material and extract information of interest. Recent efforts have shown promise in efficiently retrieving cancer-specific genomic alteration and treatment mentions from PubMed and may allow us to successfully incorporate scientific findings in precision oncology. For example, PubTator 3.0 is a tool the National Library of Medicine built to identify several entities and their relations in the literature using state-of-the-art Artificial Intelligence (AI) techniques[3]. PubTator addresses a major limitation in previous efforts that focused on processing the titles and abstracts rather than the full text of biomedical articles. The importance of this richer text is highlighted in recent efforts that have released annotated full-text corpora and/or conducted relevant Challenges[4-6].

These emerging needs inspired our exploration of automated information retrieval from biomedical literature assisted by modern NLP models and technologies. Our main focus was identifying entities essential in supporting decision-making in precision oncology, including mutations, cancer types, targeted therapies, and the relations between these entities. We evaluated the performance of the selected approaches using an existing reference standard widely used in the community, demonstrated their readiness to automate certain manual and labor-intensive processes, and discussed the remaining challenges that must be addressed to support several use cases in precision oncology more efficiently.

**Methods**

*BioRED Corpus*

We used the Biomedical Relation Extraction Dataset (BioRED), a unique resource of 600 biomedical PubMed abstracts that contain several entity types (Gene, Variant, Disease, Chemical, Species, and Cell Line) with their relations at the document level[7]. Most other annotated datasets are limited as they focus on one entity, such as tmVar[8], they do not include any relation annotations among the annotated entities, such as the BC5CDR corpus[9], or annotate relations at the sentence level, such as the DDI corpus[10]. In most real use cases in precision oncology, it is essential to evaluate the entity relations at the document level, making BioRED the ideal choice for our exploration.

We focused on four entities (Gene, Variant, Disease, and Chemical) that represent a patient's clinical-genomic phenotype and relevant drugs as well as the relations between these entities to train models that would process biomedical publications and identify genotype-driven therapies for specific cancer types. Table 1 shows the distribution of the selected entities and relations in the BioRED corpus and the training, development, and testing sets; we retained the same split in our study. Per the BioRED annotation guidelines, curators were asked to annotate each relation using one of the following nine labels: Association, Bind, Cause, Comparison, Cotreatment, Drug Interaction, Negative Correlation, and Positive Correlation. The detailed statistics for these relation types are shown in Table 2. Some curators also annotated relations between Genes and Variants as well as Variants and Variants; these relations were not described in the guidelines. Considering the sparse distribution in most relation types, we treated them all as belonging to the "Association" type, such that we have association, negative correlation and positive correlation in the end.

**Table 1**. The distribution of the selected entity types and their relations in the BioRED training, development, and test sets; the numbers in parentheses show unique entity mentions. All statistics were retrieved from the BioRED publication[7].

| Sets | Abstracts | Entity Types | | | | Relations |
|---|---|---|---|---|---|---|
| | | Gene | Variant | Disease | Chemical | |
| Training | 400 | 4430 (1141) | 890 (420) | 3646 (576) | 2853 (486) | 4178 |
| Development | 100 | 1087 (268) | 250 (135) | 982 (244) | 822 (184) | 1162 |
| Testing | 100 | 1180 (399) | 241 (137) | 917 (244) | 754 (170) | 1163 |
| Total | 600 | 6697 (1643) | 1381 (678) | 5545 (778) | 4429 (651) | 6503 |

**Table 2**. Distribution of relation types for each entity pair in the BioRED corpus. The grey-shaded cells represent relation types that do not apply to the corresponding entity pairs, per the BioRED annotation guidelines. **G**: Gene; **D**: Disease; **V**: Variant; **C**: Chemical.

| Relation Types | Entity Pairs | | | | | | |
|---|---|---|---|---|---|---|---|
| | \<G,D\> | \<G,G\> | \<G,C\> | \<D,V\> | \<C,D\> | \<C,V\> | \<C,C\> |
| Association | 1483 | 706 | 1555 | 486 | 176 | 55 | 123 |
| Bind | | 61 | 27 | | | | 1 |
| Cause | | | | | | | |
| Comparison | | | | | | | 39 |
| Conversion | | | | | | | 4 |
| Cotreatment | | | | | | | 50 |
| Drug Interaction | | | | | | | 13 |
| Negative Correlation | 79 | 128 | 298 | 15 | 446 | 10 | 174 |
| Positive Correlation | 69 | 332 | 261 | 392 | 443 | 11 | 84 |

*Natural Language Processing Solutions*

We initially evaluated several BERT models for the NER and the RE tasks. The BioBERT and BioLinkBERT were two top-performing BERT models evaluated over the Biomedical Language Understanding and Reasoning Benchmark[11, 12], and we selected them for both tasks. We also employed two LLMs, the open-source Mixtral-8x7b Instruct and the openAI's ChatGPT 4, which have already been investigated in several studies in biomedicine[13-17] and explored their performance in the NER task only. Lastly, we used PubTator 3.0, a tool built by the National Library of Medicine that, as mentioned above, identifies several entities and their relations in biomedical literature[3], including those analyzed in our study.

In the BERT exploration, we treated the NER task as a sequence labeling problem, annotating each token using the Beginning-Inside-Outside (BIO) tagging scheme. According to this scheme, the first token in an entity mention is labeled with the B-entity_type tag, subsequent tokens within the same entity are labeled with the I-entity_type tag, and all other tokens are labeled with the O tag. For example, in the sentence "Iodide transport defect (ITD) is a rare disorder", the tokens were tagged as "B-disease, I-Disease, I-Disease, B-Disease, O, O, O, O", respectively, as the ITD abbreviation was considered a new Disease mention. We then trained our BERT models using the training set and predicted each token based on the highest probability of this token. For the RE task, we pursued the identification of relations at the document level. However, BERT models can only accept a limited number of tokens, making it impossible to process a full abstract in a single pass. We, therefore, annotated each relation pair if the two entities appeared within nearby sentences (within a 3-sentence window: one before and one after) and used entity markers to highlight the entities. Then, we treated the RE task as a sequence classification problem using the relation type between the two entities as the label to train the model. For example, we converted the sentence "Crocin improves lipid dysregulation in subacute diazinon exposure through ERK1/2 pathway" into "Crocin improves <e1> lipid </e1> dysregulation in subacute <e2> diazinon </e2> exposure through ERK1/2 pathways", linked the two entities, assigned the Association label and trained the model to learn this relation. As mentioned above, we did not use the relation types described in the BioRED guidelines and found in the annotated corpus. Instead, we categorized bind, comparison, conversion, cotreatment and drug interaction relations under the broader Association type. This process was trained  using 1 Nvidia V100 GPU with a 32GB RAM cluster.

---

**Instruction**
Each of the following examples include the title, abstract, and annotations for PubMed records.

18791947|t|A case of Bernard-Soulier Syndrome due to a homozygous four bases deletion (TGAG) of GPIbalpha gene: lack of GPIbalpha but absence of bleeding.
18791947|a|More than 20 DNA mutations with different inheritance pattern have been described in patients with Bernard-Soulier Syndrome[...]

| 18791947 | 10 | 34 | Bernard-Soulier Syndrome | DiseaseOrPhenotypicFeature |
|---|---|---|---|---|
| 18791947 | 66 | 81 | deletion (TGAG) SequenceVariant | |
| 18791947 | 85 | 94 | GPIbalpha | GeneOrGeneProduct |
| 18791947 | 109 | 118 | GPIbalpha | GeneOrGeneProduct |
| 18791947 | 134 | 142 | bleeding DiseaseOrPhenotypicFeature | |
| 18791947 | 243 | 267 | Bernard-Soulier Syndrome | DiseaseOrPhenotypicFeature |

[...]

17495183|t|Tenomodulin is associated with obesity and diabetes risk: the Finnish diabetes prevention study.
17495183|a|We recently showed that long-term weight reduction changes the gene expression profile of adipose tissue in overweight individuals[...]

| 17495183 | 0 | 11 | Tenomodulin | GeneOrGeneProduct |
|---|---|---|---|---|
| 17495183 | 31 | 38 | obesity | DiseaseOrPhenotypicFeature |
| 17495183 | 43 | 51 | diabetes DiseaseOrPhenotypicFeature | |
| 17495183 | 70 | 78 | diabetes DiseaseOrPhenotypicFeature | |
| 17495183 | 205 | 215 | overweight | DiseaseOrPhenotypicFeature |

[...]

**Question**
Process the title and abstract of the new PubMed record and return the annotations in new lines.
<PMID|t|....>
<PMID|a|....>

---

**Figure 1**. Few-shot prompt engineering in the LLM exploration. The Instruction section contained five PubMed records (titles and abstracts) from the BioRed training set with the annotated entities and their exact location in the text. These examples instructed the LLMs to process the new PubMed records provided in the Question section and generate the annotated entities with their location in the text.

In the LLM analysis, we randomly selected five abstracts to use as instructional training for the model from the set of BioRed training abstracts that contained all four entities of interest (N=62). As shown in Figure 1, these five abstracts, along with their lists of annotated entities, were fed into the prompt before asking for labeled output on a new abstract provided in the same format. In the case of the Mixtral-8x7b Instruct model, we used the Microsoft Azure Databricks environment to run the analysis by setting the sampling temperature to 0.5 to decrease randomness and the maximum number of generated tokens to 2000 to receive the complete output for long abstracts contained in the BioRED testing set. For the ChatGPT4 model, we directly used the chat interface from openAI to deliver the instructions and examples. In both cases, we asked the LLM to annotate 5 abstracts before resetting the session (to minimize hallucinations observed with longer inputs) and provided the same instructions each time.

We subsequently used the PubTator 3.0 online user interface to retrieve the entities and relations for all abstracts in the BioRED testing set. At the time of retrieving this information, the PubTator 3.0 Application Programming Interface (API) was rather unstable, making it easier to obtain this data from the user interface.

The performance of all models was evaluated using the standard metrics of recall, precision, and F1-score. For the NER task, we calculated the strict version of these metrics based on the exact matches (same span of text, annotation label, and boundary) between the annotated entities in the reference BioRED testing set and the model output. In the case of the two LLMs, we also used the relaxed versions of the selected metrics by allowing a boundary overlap (at least one token) between the entity mentions as long as they were annotated with the same label. In the RE task, only the strict recall, precision, and F1 were calculated.

*Use Case*
Besides published literature in PubMed, other important resources contain information on variant actionability and pathogenicity, such as the PubMed LitVar[18], the Variant Interpretation for Cancer Consortium meta-knowledgebase[19], and the Clinical Interpretation for Variants in Cancer knowledge base[20]. A commonly used and major resource in precision oncology is OncoKB, a comprehensive knowledge base that contains detailed information about genomic alterations in cancer and clinical actionability based on pre-defined levels of evidence[21, 22]. The information in OncoKB comes from human curators who process several data sources, including scientific literature, in a comprehensive process to identify relations between key entities (genes, alterations, cancer types, and drugs) and include them in publicly available summarized tables to assist precision oncology experts in their decision-making in MTB and other settings. This labor-intensive process might benefit from applying automated tools and models, such as the ones explored in our work, that may expedite the collection of all scientific findings on the above relations, quickly synthesize the corresponding knowledge, and make it immediately available to the community. We, therefore, sought to evaluate the efficiency of the two BERT models and PubTator 3.0 to detect known relations and findings already described in OncoKB for a specific mutation.

We selected the PIK3CA E545K mutation as a representative example of an actionable mutation associated with FDA-approved targeted therapy and pulled the corresponding webpage in OncoKB on February 23, 2024. This webpage contains a detailed description of this mutation, the cancers it has been found in, the targeted drugs, the levels of evidence, as defined in OncoKB, and other supporting information; these are manually retrieved from twelve PubMed-indexed journal papers and three conference abstracts. OncoKB further summarizes the mutation-cancer-drug relations in a table under the "Therapeutic" tab, which also cites four (of the twelve) and two (of the three) above journal papers and conference abstracts, respectively. We pulled the four papers and abstracts and processed them with BioBERT and BioLinkBERT to retrieve all relations between the four entities (Gene, Variant, Disease, and Chemical) of interest. In parallel, we queried PubTator 3.0 using the four PubMed IDs to identify the relations between the entities of interest. Subsequently, the medical experts (MN, MF, and VA) of this study evaluated the output from the BERT models and PubTator and compared it with the information listed on OncoKB's summarized table for PIK3CA E545K, treating it as the reference standard. This analysis helped us determine whether these tools accurately captured the key findings from the corresponding sources without any human intervention.

**Results**
*Model Performance*
As shown in Figure 2, PubTator 3.0 achieved the highest performance in the NER task with an F1-score of 0.9 or higher across all entity types, followed by BioBERT, which efficiently retrieved Genes and Drugs with balanced recall and precision between 0.86 and 0.89. When comparing the results between BioBERT and PubTator, we found that many of BioBERT's incorrect predictions stem from its tendency to rely heavily on the surrounding context of a mention to determine its label. This often results in inconsistencies, where the same mention is assigned different entity types, which is generally not the case in medical texts. For example, in the sentence "Congenital long QT syndrome (LQTS) with in utero onset of the rhythm disturbances...," BioBERT correctly identified LQTS as a disease. However, in another sentence, "A novel spontaneous LQTS-3 mutation was identified in the...," LQTS was incorrectly predicted as a gene. We hypothesize that this behavior arises because the BERT model, trained for masked language prediction tasks, tends to prioritize context words over medical accuracy. BioLinkBERT was less efficient than BioBERT and PubTator 3.0 in supporting the NER task, with the highest recall for the Gene entity at 0.71 and other metrics between 0.55 and 0.66 for the remaining entities. Both LLMs performed poorly overall, although ChatGPT 4 demonstrated some promise, particularly in relaxed recall for the Disease and Gene entities at 0.81 and 0.78, respectively.

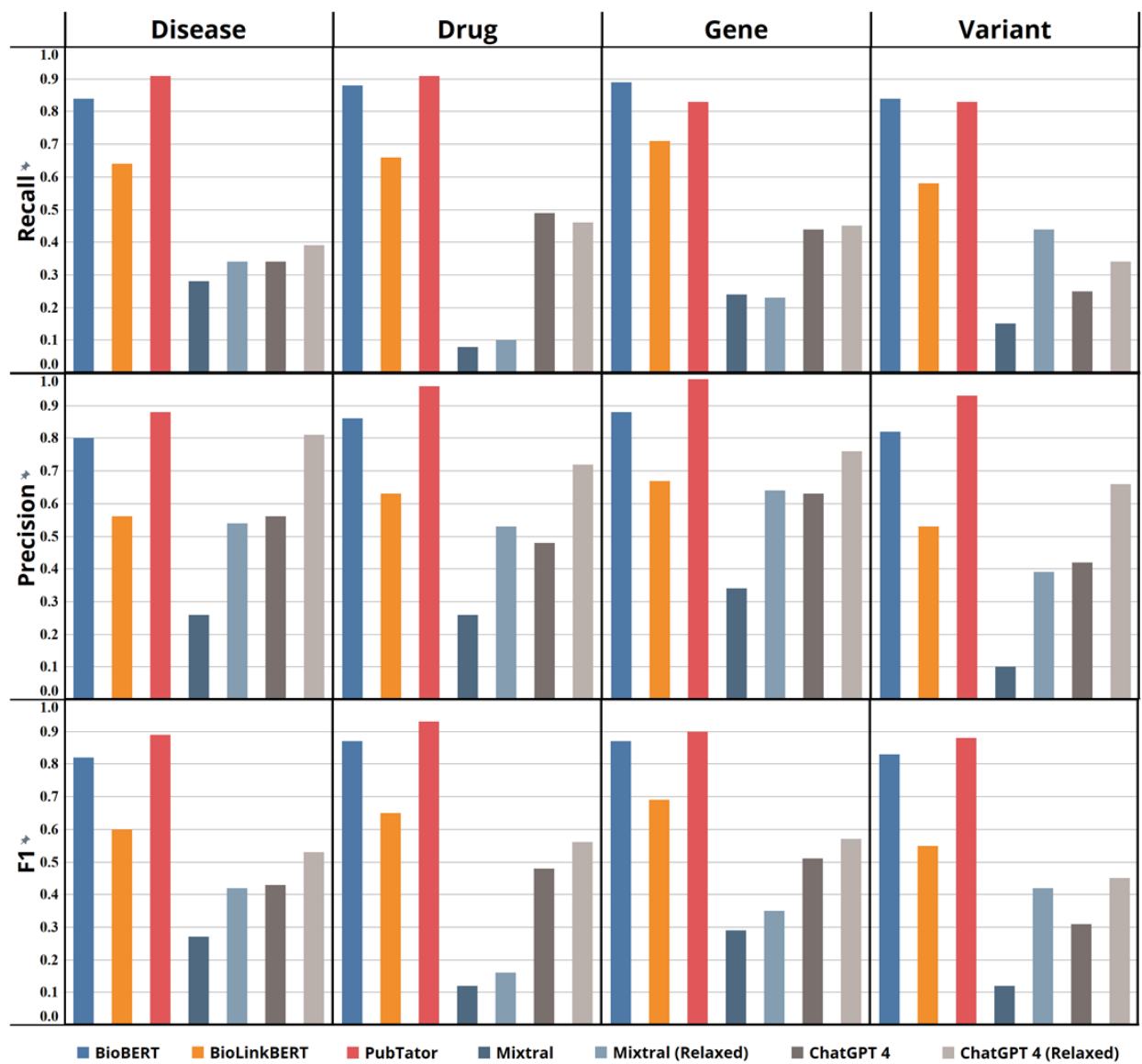

**Figure 2**. The performance of BioBERT, BioLinkBERT, PubTator 3.0 (PubTator), Mixtral-8x7b Instruct (Mixtral), and ChatGPT 4 in the NER task using the strict metrics for all solutions and the relaxed metrics for the two LLMs.

The RE task was particularly challenging. BioBERT outperformed both BioLinkBERT and PubTator, excelling in identifying Chemical-Disease relations with a recall of 0.77. However, its performance on other relations was less efficient with F1-score around 0.60, and it missed half of the Chemical-Chemical relations with recall at 0.50. The best F1-score for the BioLinkBERT was 0.47 for the Gene-Gene relation, while the lowest was 0.24 for the Chemical-Variant relations, which are essential in precision oncology. Quite interestingly, although PubTator 3.0 identified nearly all entity mentions, it struggled with relation extraction, achieving the best recall at 0.36 for the Chemical-Disease relations and F1-score ranging from 0.12 to 0.37 for other relations. Figure 3 summarizes all findings for the RE task.

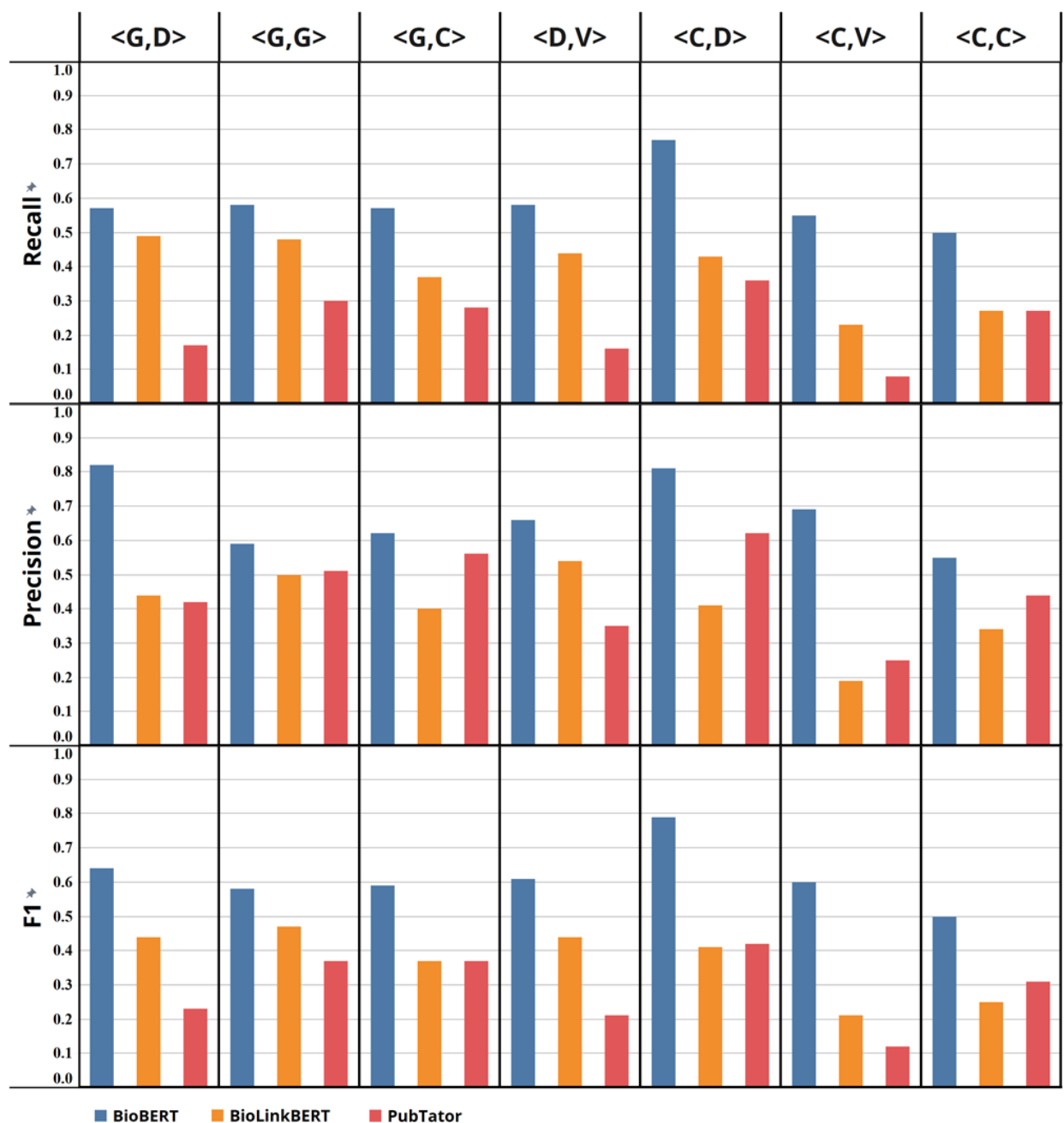

**Figure 3**. The performance of BioBERT, BioLinkBERT, and PubTator 3.0 in the RE task. **G**: Gene; **D**: Disease; **V**: Variant; **C**: Chemical.

### Use Case

As mentioned above, four journal papers and two conference abstracts are cited in OncoKB's summarized table under the "Therapeutic" tab for the PIK3CA E545K mutation. One of the medical experts retrieved all relations listed in OncoKB's summarized table and evaluated whether they were included in the BioBERT, BioLinkBERT, and PubTator 3.0 outputs and discussed the findings with the other two medical experts. Only BioBERT identified at least one relation in all papers and abstracts, while BioLinkBERT did not output any relations. As shown in Table 3, BioBERT outperformed the other two solutions by identifying 55% of all findings (21 out of 38), which, on average, did not deviate much from the corresponding numbers in the BioRED testing set. On the other hand, BioLinkBERT and PubTator 3.0 performed poorly, recognizing only 0 and 9 findings, respectively. It should be noted, though, that

PubTator 3.0 provides only relations found at the abstract and not full-text level, which partially explains its low performance.

**Table 3**. The ability of the BERT models and PubTator 3.0 to support knowledge synthesis in biomedical literature. BioBERT identified several entities in the NER task (shown in blue) but not their corresponding relations. It could not find some of the variants because the corresponding papers cited in OncoKB did not include them (C420R, Q546E, and Q546E, shown in red) or contained the altered codon without further specifying the effect of the mutation on the amino acid sequence (E545 and H1047, shown in brown). P: Paper; A: Abstract; **BB**: BioBERT; **BLB**: BioLinkBERT; **PT**: PubTator 3.0.

| A/A | OncoKB Finding | Source | BB | BLB | PT |
|---|---|---|---|---|---|
| 1 | **C420R** and **breast cancer** | P | | | |
| 2 | **Q546E** and **breast cancer** | P | | | |
| 3 | **Q546R** and **breast cancer** | P | | | |
| 4 | **E545**A and **breast cancer** | P | | | |
| 5 | **E545**D and **breast cancer** | P | | | |
| 6 | **E545**G and **breast cancer** | P | | | |
| 7 | **E545**G and **breast cancer** | P | | | |
| 8 | **H1047**L and **breast cancer** | P | | | |
| 9 | **H1047**Y and **breast cancer** | P | | | |
| 10 | E542K and breast cancer | P | ✓ | | |
| 11 | H1047R and breast cancer | P | ✓ | | |
| 12 | PIK3CA E545K and Breast Cancer | P | ✓ | | ✓ |
| 13 | Breast Cancer and Fulvestrant | P | ✓ | | ✓ |
| 14 | Breast Cancer and Alpelisib | P | ✓ | | ✓ |
| 15 | Breast Cancer and Alpelisib + Fulvestrant | P | ✓ | | |
| 16 | PIK3CA and Alpelisib + Fulvestrant | P | ✓ | | |
| 17 | Alpelisib + Fulvestrant | P | ✓ | | ✓ |
| 18 | PIK3CA and Fulvestrant | P | ✓ | | ✓ |
| 19 | PIK3CA and Alpelisib | P | ✓ | | ✓ |
| 20 | Breast Cancer and Capivasertib | P | ✓ | | ✓ |
| 21 | Breast Cancer and Fulvestrant | P | ✓ | | ✓ |
| 22 | PIK3CA and Fulvestrant | P | ✓ | | |
| 23 | Capivasertib and Fulvestrant | P | ✓ | | ✓ |
| 24 | **E545K** and **breast cancer** | P | | | |
| 25 | **Breast Cancer** and **Capivasertib + Fulvestrant** | P | | | |
| 26 | **PIK3CA** and **Capivasertib + Fulvestrant** | P | | | |
| 27 | **PIK3CA** and **Capivasertib** | P | | | |
| 28 | All Solid **Tumors** and **RLY-2608** | A | | | |
| 29 | **Breast Cancer** and **RLY-2608 + Fulvestrant** | A | | | |
| 30 | **Breast Cancer** and **Fulvestrant** | A | | | |
| 31 | **PIK3CA** (*E545K*) and **RLY-2608 + Fulvestrant** | A | | | |
| 32 | PIK3CA and RLY-2608 | A | ✓ | | |
| 33 | PIK3CA and solid tumors | A | ✓ | | |
| 34 | Breast Cancer and RLY-2608 | A | ✓ | | |
| 35 | RLY-2608 and Fulvestrant | A | ✓ | | |
| 36 | PIK3CA (*E545K*) and RLY-2608 | A | ✓ | | |
| 37 | PIK3CA (*E545K*) and Fulvestrant | A | ✓ | | |
| 38 | PIK3CA and E545K | A | ✓ | | |

Because BioBERT was the best-performing solution in the earlier RE task, we took a deeper dive to evaluate the missed relations on this task and discovered that, in almost half of them (8 out of 17), the model pulled all participating entities correctly but could not associate them (Table 3, lines 24-31). In the remaining relations, the model identified the cancer type, i.e., "breast cancer", but not the associated variant. , This phenomenon was attributed to two reasons: the corresponding papers cited in OncoKB either did not include them (C420R, Q546E, and Q546E; Table 3, lines 1-3) or reported the altered codon without further specifying the amino acid change (E545 and H1047; Table 3, lines 4-9). It appears that the OncoKB human curators used their expertise to interpret the published findings, making inferences from relevant knowledge acquired in separate explorations without providing complete references on the PIK3CA E545K webpage. None of these could have been captured by the solutions explored in our study.

## Discussion

We explored the ability of selected NLP solutions to efficiently retrieve specific entities (Gene, Variant, Disease, and Chemical) and their relationships from biomedical literature. Additionally, we assessed how these solutions might contribute to decision-making in precision oncology through a particular use case. In the NER task, PubTator 3.0 was the best-performing solution (F1-score close to or above 0.9), followed by BioBERT (F1-score between 0.82 and 0.89), while BioLinkBERT and the two LLMs demonstrated average and poor performance, respectively. In the RE task, BioBERT outperformed BioLinkBERT and PubTator 3.0, but did not reach the required level of performance for routine use, missing several relations in the evaluation with the BioRED testing set and the OncoKB use case. However, it captured all individual entities included in retrieved or missed relations and reported in the OncoKB-cited papers.

Our study has three main limitations. First, the selected BioRED corpus was not originally created with a focus on precision oncology, e.g., the Disease entity did not solely represent cancer types, and the annotated relations did not fully capture all potential associations between the Chemical and the other entities, which is a critical factor in the selection of targeted therapies. Dedicated resources might better support information retrieval and knowledge synthesis from biomedical literature. Second, none of the explored solutions could make inferences from other sections within the same full-text papers or other sources in general, as depicted in the OncoKB use case. This limitation applies to detecting relations at the document level and not simply at the section level, which partly occurred in our approach. It also refers to linking several documents and recognizing entity relations across them. Generative AI may likely bridge this gap, which may represent an additional challenge, as we elected to use LLMs for annotation rather than knowledge generation purposes.

The presented level of performance in the RE task cannot lead to operationalizing any of the solutions evaluated in our work. We would argue, though, that the domain has not solved this problem yet, especially in precision medicine and oncology. A representative example is the recently developed BioREX model, the core RE engine in PubTator 3.0, that considerably improved RE in several relation types (average F1-score 0.79) by applying deep learning to heterogeneous datasets (BioRED was one of these datasets)[23]. However, none of these relations contained the Variant entity, although annotated in several datasets[24, 25], which is significant for characterizing a patient's genomic profile and selecting targeted therapies in precision oncology. BioREX's inability to efficiently capture this relation type was demonstrated in PubTator's evaluation in the OncoKB use case. In that sense, although PubTator performed best in the NER task, our BioBERT model offered a more efficient approach by nearing PubTator in the NER and outperforming it in the RE task. It should be clarified, though, that none of these approaches is mature enough to solve the RE problem.

Acknowledging the limitations of existing approaches in accurately detecting certain relations in biomedical literature and clinical texts, we suggest pursuing other strategies in the precision oncology context. Traditional NLP approaches would require annotating new corpora or combining and refining existing labeled datasets and (re)training some of the best-performing state-of-the-art models to improve performance. One might also argue that the strengths of the BERT models have not been fully explored in precision oncology, and we would probably agree with this statement. For example, our BioBERT model, as part of an NLP ensemble pipeline, could accept PubTator's NER output and efficiently detect the relations in biomedical literature. This approach was not examined in our work but could be investigated in one of our next steps. On the other hand, the research community is trying to move from training models to "zero-shot" learning frameworks[26] that promise less labor-intensive processes and efficient implementations in end-to-end systems utilizing LLMs. The current work demonstrated that the selected LLMs could not accurately support the NER task traditional NLP approaches have (nearly) solved, further suggesting potential major challenges for these models in the more complex RE task, which we investigated in a separate analysis and found them performing very poorly (data not shown).

The application of generative AI to several tasks in biomedicine and precision oncology is inevitable, and we will be seeing numerous studies in this area over the next several years. Our limited analysis of the two LLMs represents a pilot study and, as such, precludes firm conclusions on the use of these methodologies in clinical decision-making. It is paramount to accurately collect the requirements and expectations from the end users before determining the next steps and calibrating these approaches. In precision oncology specifically, several sources are evaluated to make a clinical decision, and biomedical literature is only one of them. Moreover, knowledge synthesis must be coupled with several other processes, including, but not limited to, accurate computable phenotyping and efficient data integration.

**Conclusion**

We explored several NLP solutions to automatically extract, synthesize and characterize scientific knowledge from biomedical literature that might support clinical decision-making in the context of identifying genotype-driven therapies for cancer patients. Identifying the relations between key entities was the most challenging task in our analysis, as shown in the comparisons with the reference standard and knowledge included in the OncoKB resource. The latter evaluation was very informative and demonstrated that one of the models (BioBERT) successfully identified all entity mentions found in the OncoKB-cited publications and 55% of the relations listed by the human curators. Future research must deliver efficient systems that will accurately process the compendium of information and process knowledge to support decision-making in precision oncology.

**Acknowledgments**

This study was supported by the National Cancer Institute as part of two research awards (U01CA274631 and P30CA006973). The contents are those of the authors and do not necessarily represent the official views of, nor an endorsement, by NCI or the U.S. Government.